# Considerations upon the Machine Learning Technologies

**Assist.Prof. Alin Munteanu, PhD**
**Instructor Cristina Ofelia Sofran, PhD Candidate**
„Tibiscus" University of Timișoara, Romania

REZUMAT. Inteligența artificială oferă tehnici sau metode evoluate prin care probleme din variate domenii își pot găsi o rezolvare optimă. Tehnologiile Machine Learning se referă la domeniul inteligenței artificiale ce are ca scop dezvoltarea de tehnici ce permit calculatoarelor „învățarea". Unele sisteme bazate pe tehnologii Machine Learning tind să elimine necesitatea intuiției umane, în timp ce altele adoptă o abordare colaborativă între om și mașină.

## 1. Introduction to Machine Learning

Machine learning techniques define an artificial intelligence domain which has a main goal to develop techniques allowing the computers to "learn". To be more specific, Machine learning is a means to create programs through analyzing data sets. Machine learning technologies and statistics are connected since they both have data analyzing as a study objective, but unlike statistics, machine learning focuses on algorithmic complexity of computational implementations. These machine learning technologies have a large area of applicability, including search engines, setting medical diagnostics, detecting fraud with an electric card, stock analysis, speech and writing recognition, gaming and robot kinetics.

Some machine learning based techniques tend to eliminate the need of human intuition for analyzing the data, while other adopt a collaborative approach between man and machine. Human intuition cannot be completely removed since the system designer has to specify the mode of representing data and the mechanism used to search the data. Machine learning is similar to an attempt of automatize the parts of a scientific procedure.

Machine learning refers to modifications within the systems executing various tasks related to the artificial intelligence domain, tasks involving recognition, diagnostic, planning, robot control, prevision, which can only be completely defined through examples, by supplying the input data and the expected results. It is desired for the results to be able to be deducted given that an input data series exists, but there is no well-defined input-output function, only by approximating the implicitly relations. It happens many times that the correlations and links are "hidden" into the huge amount of data, but with the help of machine learning technologies, these can be extracted.

There are often designed systems that do not work efficiently in the area they are used, because some particularities of the work manner could not have been well specified when they were built, so the Machine learning methods come in handy in these cases. Information can be various and create new knowledge flows, which could cause the reimplementation of the artificial intelligence systems, but given that this is not a practical solution, it appears that the machine learning technologies could handle these situations well.

In the following we shall present and analyze some of the most representative machine learning technologies.

## 2. Machine Learning Technologies

*A. Statistical Learning*

Given a vector $X$ as a variable which distribution probability for a category $c_1$, $p(X|c_1)$ is different from the distribution probability for another category $c_2$, $p(X|c_2)$. With the given the model $X$, the usage of statistical techniques would allow resolving the distribution which it derives from. For developing a decisional model we need to know the results determined by the two types of errors that can occur, information retained using a "loss function": $\lambda(i|j)$, where $i, j = 1,2$ and $\lambda(i|j)$ is in fact the occurring loss in the case of choosing the decision from the *i* category, when actually the decision is from the *j* category. Considering the same $X$ model, if the decision would be from the *i* category, the estimated loss value would have to be:

$$L_X(i) = \lambda(i\,|\,c_1)p(c_1\,|\,X) + \lambda(i\,|\,c_2)p(c_2\,|\,X) \qquad (1)$$

where $p(j\,|\,X)$ is the probability of model $X$ should correspond to the $j$ category. The decision would be favorable for the case when $X$ belongs to the $c_1$ category if $L_X(c_1) \leq L_X(c_2)$, and it belongs to the $c_2$ category otherwise.

In general, if we defined $k(i\,|\,j)$ as $\lambda(i\,|\,j)p(j)$, then the decision rules are taking the following form: it will be decided for the 1st category if

$$k(c_1\,|\,c_2)p(X\,|\,c_2) \leq k(c_2\,|\,c_1)p(X\,|\,c_1) \qquad (2)$$

*B. Decision trees*

The decision trees represent a technique that can apply for classification as well as for prediction, the result having the shape of a tree design that reveals a hierarchy of logical rules automatically defined by querying a data base containing examples. These examples are complex records with many attributes. The rules derive from the more and more detailed subdivision of the assembly of examples, depending on the attributes content.

Some outstanding types of decision trees which are successfully utilized in learning issues are: *ID3* and its more advanced version, *C4.5*, suggested by the famous Australian professor Quinlan, and *CART*, proposed by the well-known Breiman. The CART method (Classification and Regression Trees) starts with solving the independent variable which values allow the best subdivision. Given this purpose, a diversity index of the assembly of records is being computed from the perspective of an attribute. The algorithm is defined by parsing each independent attribute one by one and computing the decrease factor of the diversity the subdivision made using this method could bring. The variable maintained as a separation criterion is the one that will lead to the best result.

$$I_G(i) = 1 - \sum_{j=1}^{m} f(i,j)^2 \qquad (3)$$

where $f(i,j)$ represents the frequency of the value *j* within the node *i*.

Unlike the CART, C4.5 can generate binary trees only. In the informatics version, C4.5 has the ability to automatically generate rules. Beginning with the full set, generated directly from the tree, the application attempts to make a generalization with the purpose of reducing the number of rules. For this, for each rule, some conditions will be eliminated and checks are being performed to verify the amount this maneuver increases the error rate. A series of other transformations can be operated for this goal, as so eventually the number of rules can reach a value below the number of leaf nodes.

$$I_E(i) = -\sum_{j=1}^{m} f(i,j) \log f(i,j) \qquad (4)$$

*C. Reinforcement learning*

Reinforcement learning is defined by learning a way to act as so the rewards can be maximized. The subject who needs to be taught isn't told which path he must take, like in most of the machine learning techniques, instead he must find out on his own which action will help him acquire the most efficient reward by trying each action out. In the most complex of the cases the actions taken will not just affect the immediate rewards, but also future rewards. Thus, the reinforcement learning technique can be distinguished through two main characteristics: the one of tryout – failure and the one of delayed rewarding. Reinforcement learning will not be defined through describing the particularities of the teaching methods, but through exposing the teaching problem. Any proper procedure for solving the problem is considered to be a *reinforcement learning* procedure. The basic idea is that all aspects of the real problem must be well understood with the assistance of a teaching agent which is capable of perceiving the domain and of taking measures which can affect the problem state.

A reinforcement learning example is the so-called *Q learning* which allows an agent to learn some rules from an arbitrary environment. The application $\pi^* : S \to A$ is difficult to assimilate for no examples of the form $\langle s, a \rangle$ are being produced, although the information available to the teacher is actually sequences of immediate rewards of the form: $r(s_i, a_i)$ where *i = 0, 1, 2,...* . The teaching agent can attempt to learn an evaluation application $V^*$ and prefer a certain state $s_1$ to another state $s_2$: $V^*(s_1) > V^*(s_2)$. The optimal action regarding a state s is one that maximizes the sum of immediate rewards $r(s,a)$ and the value of the application $V^*$ for the next state in line:

$$\pi^*(s) = \arg\max[r(s,a) + \gamma V^*(\delta(s,a))] \qquad (5)$$

As it can be deduced, the teaching agent can act in the optimal manner by assimilating the application $V^*$, considering that the reward acquired by the r function and the $\delta$ function of state transition are well-determined.

### D. Explanation Based Learning

*EBL* is very much like a process where the implicit knowledge is converted in an explicit one. This type of learning begins from a particular example, after which an explanation is generated in order to be able to use this in other situations as well. The addition of a rule is accompanied by a drop in the level of depth, but also by a raise in the number of formulas in the theoretical area. In practice, additional rules are relevant only for some questions in matter, but overall it is uncertain that the new rules really prove themselves useful. Generally, though, the EBL methods have been applied in a diversity of domains and usually lead to good results.

### E. Deductive Learning

A deductive system is one that obtains the conclusions following the logic from an input data set. Given a logical phrase $\phi$, derived from a fact set, $\Delta$, the deduction process of $\phi$ from $\Delta$ is equivalent to the assimilation of the $\phi$ process. Still $\phi$ may not be this obvious to deduct from $\Delta$ and the process of resolving the $\phi$ phrase may be harder. Instead of having the $\phi$ phrase deducted again it is preferred to retain this one in case it is needed later on, and this process is interpreted as a learning procedure.

## Conclusions

It is considered that, within companies and organizations, the classic decision support information systems no longer raise up to the level of expectations, thus the intelligent decision support systems prove to be very useful, systems that can be built using machine learning technologies.

## References

[CS91]  S. Cooke, N. Slack - *Making Management Decisions,* 2nd Edition, Prentice Hall, 1991.

[Z+01]  D. Zaharie, F. Albescu, I. Bojan, V. Ivancenco, C. Vasilescu - *Sisteme informatice pentru asistarea deciziei (Decision support information systems)*, Editura DualTech, 2001.

[Mit97] T. Mitchell - *Machine Learning*, McGraw-Hill, 1997.